\documentclass[conference]{IEEEtran}
\IEEEoverridecommandlockouts
\usepackage{cite}
\usepackage{amsmath,amssymb,amsfonts}
\usepackage{algorithmic}
\usepackage{graphicx}
\usepackage{textcomp}
\usepackage{xcolor}

\usepackage[caption=false]{subfig}
\usepackage{color}
\usepackage{soul}

\def\BibTeX{{\rm B\kern-.05em{\sc i\kern-.025em b}\kern-.08em
    T\kern-.1667em\lower.7ex\hbox{E}\kern-.125emX}}
    \newcommand*{\affaddr}[1]{#1} % No op here. Customize it for different styles.

\newcommand*{\affmark}[1][*]{\textsuperscript{#1}}

\begin{document}

\title{Keeping the Questions Conversational: Using Structured Representations to Resolve Dependency in Conversational Question Answering
}

\author{
Munazza Zaib\affmark[1,*]\thanks{\emph{*Corresponding Author:munazza-zaib.ghori@hdr.mq.edu.au}}, Quan Z. Sheng\affmark[1],  Wei Emma Zhang\affmark[2], and Adnan Mahmood\affmark[1]\\

\affaddr{\affmark[1]School of Computing, Macquarie University, Sydney, NSW 2109, Australia }\\

\affaddr{\affmark[2]School of Computer and Mathematical Science, The University of Adelaide, Adelaide, SA 5005, Australia}\\
}

\maketitle

\begin{abstract}

Having an intelligent dialogue agent that can engage in conversational question answering (ConvQA) is now no longer limited to Sci-Fi movies only and has, in fact, turned into a reality. These intelligent agents are required to understand and correctly interpret the sequential turns provided as the \textit{context} of the given question. However, these sequential questions are sometimes left implicit and thus require the resolution of some natural language phenomena such as {\em anaphora} and {\em ellipsis}. The task of question rewriting has the potential to address the challenges of resolving dependencies amongst the contextual turns by transforming them into {\em intent-explicit questions}. Nonetheless, the solution of rewriting the implicit questions comes with some potential challenges such as resulting in verbose questions and taking \textit{conversational} aspect out of the scenario by generating the self-contained questions. In this paper, we propose a novel framework, CONVSR (\underline{CONV}QA using \underline{S}tructured 
\underline{R}epresentations) for capturing and generating intermediate representations as conversational cues to enhance the capability of the QA model to better interpret the incomplete questions. We also deliberate how the strengths of this task could be leveraged in a bid to design more engaging and more eloquent conversational agents. We test our model on the QuAC and CANARD datasets and illustrate by experimental results that our proposed framework achieves a better F1 score than the standard question rewriting model.
\end{abstract}


\section{Introduction}
Conversational question answering (ConvQA) is a relatively new paradigm and considerable task that possesses the potential to revolutionize the way humans interact with machines \cite{zaib2022conversational}. It, in fact, requires a system to answer a set of interrelated questions posed by any user \cite{choi-etal-2018-quac, reddy2019coqa, gao-etal-2018-neural-approaches}. In human conversations, these sequential questions could be \textit{implicit} and are very easy for them to understand \cite{DBLP:conf/sigir/ChristmannRW22}. However, ConvQA-based machines are expected to learn and resolve such implicit dependencies from the given context. For instance, let us consider a ConvQA session pertinent to a TV show in Table~\ref{friends}. In order to interpret and answer Q2, the system is expected to have the information about Q1 and A1. Similarly, it would be difficult for the system to find the answer to Q3 since there could be many first seasons of different series. To retrieve the correct answer, the system needs to incorporate the show name in the question.

\begin{table}[!h]\centering
\caption{An example of 
%an 
information-seeking dialogue. Red denotes the context entity, whereas, blue represents the question entities.}
{\renewcommand{\arraystretch}{1.15}}
\begin{tabular}{p{1.5cm}p{5.5cm}}

      \hline
    \hline
     \multicolumn{2}{c}{Topic: F.R.I.E.N.D.S} \\
     \hline
    \textbf{ID}  & \textbf{Conversation} \\
    \hline
    Q1  & Who played \textcolor{purple}{Monica Geller} in \textcolor{blue} {FRIENDS}? \\
    A1  & Courteney Cox.\\\hline
    Q2 & What was \textcolor{purple}{she} obsessed about?\\
    A2 & Cleaning. \\ \hline
    Q3  & Who was the \textcolor{purple}{noisy neighbor}? \\
    A3  & Larry Hankin \\ \hline
    Q4 & Release date of the \textcolor{purple}{first season}? \\
    A4 & 22 September 1994.\\
    \hline
    \hline
    \\
    \end{tabular}
      \label{friends}
      \vspace{-2.5em}
\end{table}

Furthermore, the task of question rewriting (QR) has been extensively researched upon by researchers in the information extraction community. Nevertheless, it is fairly new in the field of ConvQA and is recently introduced as an independent task in some of ConvQA models \cite{vakulenko2021question, yu2020few, kim-etal-2021-learn}. 
%It 
Simply put, QR refers to the task of reformulating the given question by adding missing information or resolving co-references. This process generates a stand-alone %and independent 
question by extracting it out of the conversational context \cite{raposo2022question}. However, this results in losing valuable cues from the conversational flow. Also, the resulting rephrased questions might be long and verbose which, in turn, results in difficulty in retrieving evidence from the given context. Furthermore, the datasets available for question rewriting in ConvQA are quite small, thereby hindering the training process of the model.

To address these particular shortcomings, we propose an ensemble model entitled, CONVSR (\underline{CONV}QA using \underline{S}tructured 
\underline{R}epresentations), which instead of rewriting an incomplete or ambiguous question generates the intermediate structured representations (SR) based on the given context and the question. These representations, comprising the context and the question entities, can ultimately be used to fill up the missing gaps and answer the question at hand. The key intuition behind such a model is that an incomplete question only needs to refer to the last few questions in order to fill in the missing gaps because the conversational flow keeps on changing \cite{DBLP:conf/iclr/HuangCY19, qu2019attentive, DBLP:conf/acl-mrqa/YehC19}. Hence, to accommodate the changing conversational flow, we propose to select \textit{k} history turns using dynamic history selection process.

Some questions would be requiring both the context and the question entities in a bid to disambiguate the current question, whereas, for some questions, only the context entities would be enough. For instance, to answer Q2 in Table~\ref{friends}, the model needs to refer back to the intermediate representations captured for Q1. In this case, the model needs to have both the context entity (FRIENDS) and the question entity (Monica Geller) to decipher \textit{`she'} in Q2. However, to answer Q4, the model only needs the context entity i.e., FRIENDS.

Hence, our proposed model consists of the following main stages:
\textit{i) Question understanding} which encompasses assessing a question based on the given context; \textit{ii)} \textit{Dynamic history selection} that conducts `hard selection' for the relevant history turns. This method attends to the previous history turns based on a semantic similarity score. If a conversational turn equals or surpasses a threshold value, then it is considered important for predicting the answer; \textit{iii) Entity generation} which works towards identifying the context and question entities (if any) from the selected history turns; and \textit{iv)} \textit{ Answer prediction} that retrieves the most relevant answer span from the given context based on the selected history turns and their respective SRs.
%Thus, 
In a nutshell, the technical contributions of this research work can be summarized as the following:
\begin{itemize}
    \item We highlight the limitations of the previous approaches and propose a framework to address these limitations. Our framework presents an alternative of question rewriting task to complete the ambiguous questions by generating intermediate structured representations. 
    \item We propose a dynamic history selection policy based on `hard history selection' to only select the relevant subset of conversational turns.
    %michael: we need to elaborate the following point
    \item We study the effect of SRs on traditional ConvQA baselines by skipping the dynamic history selection process and appending the history turns in different settings.
    \item We demonstrate by our experimental results that ConvQA models suffer from a decline in accuracy by incorporating QR task within the model, thus, proving the effectiveness of our approach.
\end{itemize}

To the best of our knowledge, our proposed research is one of the few research works that implement the task of question resolution within the ConvQA setting. The rest of this paper is organized as follows. Section~\ref{sec:relatedwork} overviews the techniques on conversational question answering and question completion. 
Section~\ref{methodology} illustrates the technical details pertinent to our proposed CONVSR model. 
Section~\ref{sec:setup} describes the experimental set up and Section~\ref{result} reports the experimental results.
Finally, Section~\ref{sec:conclusion} offers some concluding remarks. 

\section{Related Work}
\label{sec:relatedwork}
\subsection{\textbf{Conversational Question Answering}}
Recent advancements in natural language processing have led to the development of (ConvQA) systems, which aim to provide accurate and contextually relevant answers to user queries in a conversational setting. These advancements are mainly owing to the rapid progress of the pre-trained language models \cite{DBLP:conf/naacl/DevlinCLT19, 10.1145/3373017.3373028, DBLP:conf/acl/LewisLGGMLSZ20, radford2018improving, DBLP:journals/corr/abs-1907-11692} and the availability of the task-specific datasets such as QuAC \cite{choi-etal-2018-quac} and CoQA \cite{reddy2019coqa}.  These developments have taken the NLP and IR communities by storm and have resulted in several state-of-the-art models.  Many research works have  introduced novel and different strategies to tackle this challenge. 

The task of ConvQA presents several challenges to the researchers hence resulting in considerable interesting yet innovative research works over the past few years. One of the key challenges in the task of ConvQA is to incorporate the conversational history effectively so that the model can best interpret the current question accurately. Some popular strategies include prepending the conversational turns \cite{choi-etal-2018-quac, reddy2019coqa, DBLP:journals/corr/abs-1905-12848} and dynamic history selection either utilizing attention mechanism  \cite{qu2019attentive} or reward-based reinforcement learning \cite{DBLP:conf/aaai/QiuHCJQ0HZ21}. Several other works also demonstrate the effectiveness of FLOW-based mechanisms \cite{DBLP:conf/acl-mrqa/YehC19, DBLP:conf/iclr/HuangCY19,DBLP:conf/ijcai/0022WZ20} to capture the intermediate latent representations to help the answering process. We employ a dynamic history selection process to obtain question-relevant conversation history turns. Integrating non-relevant conversational turns tend to bring noise into the input provided to the model, which in turn, results in the model's performance degradation \cite{zaib2021bert, qu2019attentive, DBLP:conf/aaai/QiuHCJQ0HZ21}. The process is based on hard history selection and will be discussed in Section~\ref{methodology}. 
\begin{figure*}[!t]
\begin{center} 
\includegraphics[width=0.95\linewidth, height=2.3in]{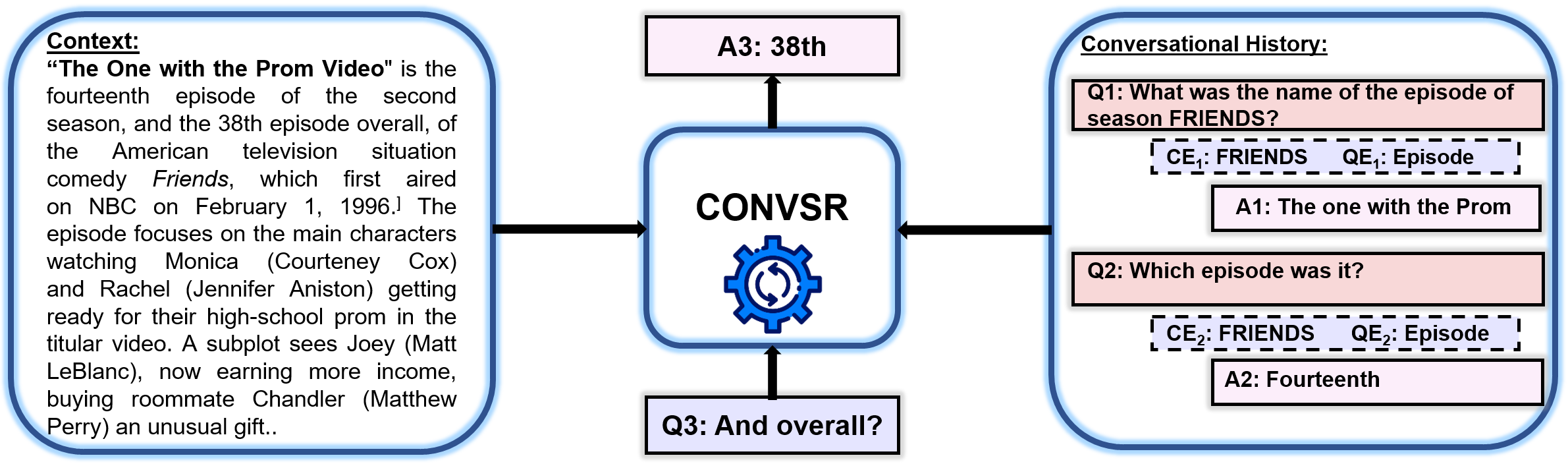}
\end{center}
\caption{An illustration of our proposed model Conversational Question Answering with Structured Representations (CONVSR). CE and QE in CONVSR denote the context entity and question entity. }
\label{fig:overall-arch}
\vspace{-3mm}
\end{figure*}
\subsection{\textbf{Question Completion}}
A popular research direction that aims to address the challenges pertinent to an incomplete or ambiguous question is question rewriting (QR). The task of QR is recently adopted in the field of ConvQA to rephrase and generate a self-contained question that can be answered from the given context \cite{raposo2022question, DBLP:conf/wsdm/VakulenkoLTA21, kim-etal-2021-learn, elgohary-etal-2019-unpack, li-etal-2022-ditch, chen2022reinforced}. However, the task of QR takes the conversational questions out of the context by transforming them into self-contained questions, which in turn, negates the whole idea of ConvQA setting \cite{DBLP:conf/sigir/ChristmannRW22}. Question resolution is another approach that adds relevant and significant terms from the previous conversation turns to fill in the missing information gaps \cite{10.1145/3397271.3401130}. These techniques are widely used to resolve co-dependencies and anaphora among the conversational history turns. 

We work on the second type of question completion technique and show in our experimental results in Section~\ref{result} that question reformulation with added valuable cues performs better than rewriting questions from the scratch. 

\section{Methodology}
\label{methodology}

\subsection{\textbf{Task Formulation}}
We assume the traditional setting of ConvQA where a user starts the conversation with a particular question or information need and the system searches the given context to provide an answer after each of the user's questions \cite{zaib2021bert}.
\begin{table}[htb!]\centering
\caption{The table shows the example highlighted in Fig~\ref{fig:overall-arch}. To answer Q3, the model will utilize context and question entities from Q2 and Q1. In this example, the SRs for Q1 and Q2 are same, thus, showing that the conversational flow until Q3 is the same. Q3 can be interpreted by the CONVSR model as \textit{`And overall?  (\textcolor{blue}{FRIENDS}, \textcolor{purple}{episode})'}}
{\renewcommand{\arraystretch}{1.15}}
\resizebox{\linewidth}{!}{
\begin{tabular}{p{1.5cm}p{5.5cm}}

     \multicolumn{2}{c}{   SR = (\textcolor{blue}{context entity} $|$ \textcolor{purple}{question entity})} \\
    
    \hline
    Q1  & What was the name of the episode of season FRIENDS? \\
    A1 & The one with the prom video.\\
    \hline
    $SR_1$ & (\textcolor{blue}{FRIENDS} $|$ \textcolor{purple}{episode})\\
    Q2 & Which episode was it?\\
    A2 & Fourteenth \\
    \hline
    $SR_1$ & (\textcolor{blue}{FRIENDS} $|$ \textcolor{purple}{episode})\\
    $SR_2$ & (\textcolor{blue}{FRIENDS} $|$ \textcolor{purple}{episode})\\
    Q3  & And overall?\\
    A3 & 38th\\
    \hline    
    \\
    \end{tabular}}
    
      \label{friends1}
      \vspace{-3em}
\end{table}

The follow-up questions may be incomplete or ambiguous requiring more context to be interpreted by the model (example: `What was she obsessed about?').
The task of CONVSR is to capture the context entities and question entities from the previous relevant conversation turns and utilize them as additional cues to answer incomplete questions. The term \textit{context entity} corresponds to an entity mentioned in the previous conversational context and the term \textit{question entity} corresponds to the entities given in the previous questions. Essentially, an SR for any given question can be represented as shown in Table~\ref{friends1}.

More formally, given a conversational context \textit{C}, previous history turns \textit{H} and a potentially ambiguous or incomplete question \textit{Q} which may need the understanding of the previous conversation turns, the task of CONVSR is to first select the relevant history turns $H^{'}$ and then capture the structured representations \textit{SR} in the form of context entity \textit{CE} and question entity \textit{QE}. These SRs are then infused into the ConvQA model to be utilized to generate the correct answer \textit{A}. An illustration of our proposed model CONVSR is depicted in Fig.~\ref{fig:overall-arch}.

\subsection{\textbf{Pipeline Approach}}
Over the past few years, a number of research works \cite{kim-etal-2021-learn, vakulenko2021question,raposo2022question,vakulenko-etal-2020-wrong,ishii-etal-2022-integrating} have envisaged various models to tackle the complexity of ConvQA task by decomposing it into QR and QA subtasks.  Question rewriting, being the initial sub task, generates self-contained question by rewriting the given incomplete question from the scratch. Different approaches are in use to generate these rewrites such as language models \cite{raposo2022question, vakulenko-etal-2020-wrong, ishii-etal-2022-integrating, DBLP:journals/corr/abs-2004-01909} and neural networks \cite{vakulenko2021question}.
%etc. 
The QR models are trained on a recently introduced CANARD \cite{elgohary-etal-2019-unpack} dataset, which is based on QuAC's \cite{choi-etal-2018-quac} original questions and their respective rewrites. The dataset has around 40K question pairs generated by human annotators. 

\begin{figure*}
\centering
\subfloat[Pipeline approach \label{qr}]{\includegraphics[width=.4\linewidth]{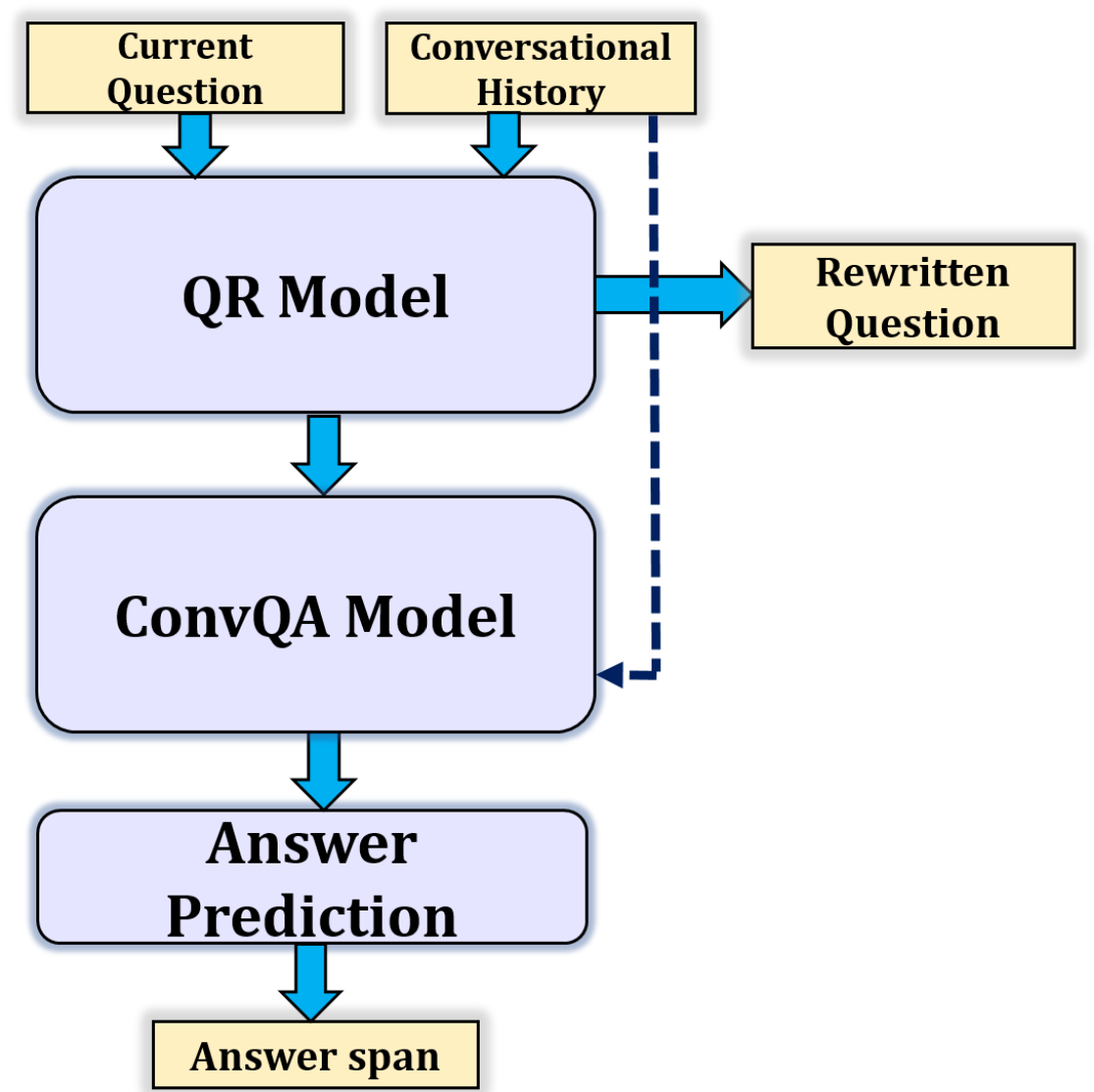}} \qquad 
\subfloat[Our proposed approach \label{sr}]{\includegraphics[width=.4\linewidth]{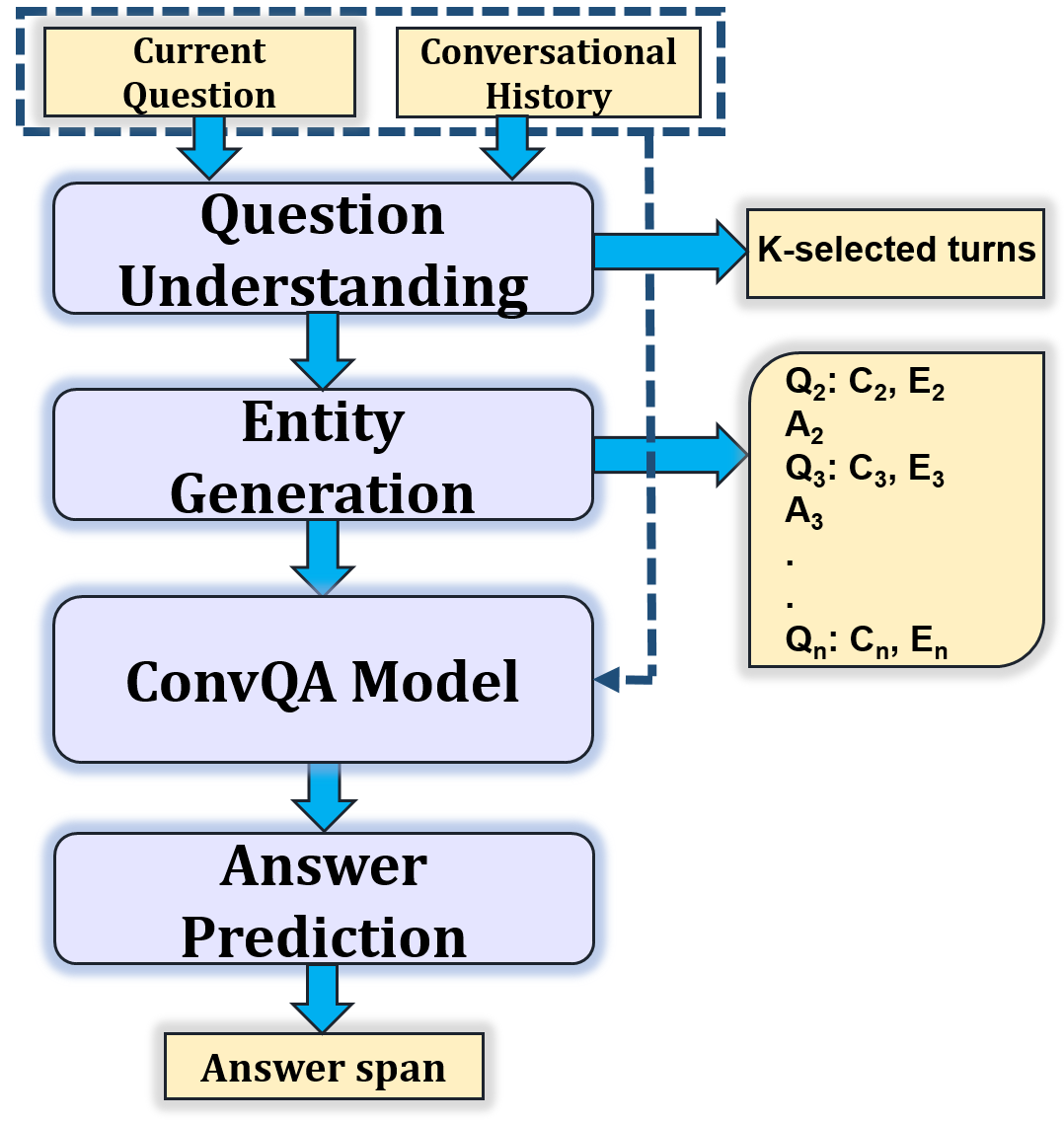}} 
\caption{In the pipeline approach, a context-independent question is generated by the QR model which serves as an input to the ConvQA model to predict an answer span. In our approach, first relevant history turns are selected and SRs are generated for them. The SRs, current question and history turns form an input for the QA model to find the accurate answer span.}
\vspace{-1em}
\end{figure*}

Following \cite{DBLP:journals/corr/abs-2004-01909}, we adopt GPT-2 \cite{radford2018improving} to train the QR model. In the training process, we provide the conversational turns and the current question as the inputs and the model generates a context-independent rewrite that is to be answered without taking the conversational history into consideration. 
Once the rewrites are generated, the next sub-task is of the QA module to find a relevant answer from the given context. Since it is assumed that all the co-references and anaphoras have been resolved in the QR task, most research works employ a traditional QA model instead of the ConvQA framework to answer the current question. However, we utilize conversational history along with SRs in our proposed model, therefore, for a fair comparison we also utilize conversational history along with the rewritten question as input to the QA model. We put together the process of predicting an answer as:
\begin{equation}
P =  (a_i,|q_i,C, H) \approx P^{qr}(q_{i}^{'}|q_i,H) \cdot P^{qa}(a_i|q^{'},C, H)
\end{equation}
where $P^{qr} (\cdot)$ and $P^{qa} (\cdot)$ are the likelihood functions of the two sub-task models, respectively. $q^{'}$ represents the rewritten question by the QR model and it serves as an input to the QA model along with the given context and history turns. The pipeline model is shown in Fig.~\ref{qr}. The dotted line represents that conversational history forms an input to the ConvQA model along with the rewritten question.

The primary limitation of using this approach is that the QA model never gets to be trained on the user's actual questions, and tends to loose the understanding of the conversational context. Also, the input of a QA model is highly dependent on the output of the QR model, which increases the chances of QA model being suffered by error propagation from QR model.

\subsection{\textbf{The CONVSR Model}}
Follow-up questions in a conversation are usually incomplete and require explicit information to identify the question's intent. Thus, the key challenge pertinent to ConvQA involves understanding the conversational flow to derive the structured representations to aid the answering process of the user's information need. Instead of following the conventional approach of question rewriting, we rather aim to capture and extract intermediate SRs to keep the task conversational. These SRs serve as an aid to an incomplete question by filling in the gap by adding more context using context entity and/or by resolving co-references using question entity. However, generating SR for all the previous questions would bring in the noise data (irrelevant information), which tends to decrease the model's performance as proved in \cite{zaib2021bert, 10.1145/3331184.3331341}. Also, the follow-up questions usually take their context and question entities from the previous immediate turns. The dialogue behavior of \textit{topic shift} and \textit{topic return} requires the explicit mention of the new entity and the context. 

We utilize a pre-trained seq2seq language model, BART \cite{DBLP:conf/acl/LewisLGGMLSZ20} to perform history selection and generate SR entities. It consists of both an encoder and a decoder. The model is best utilized when the information is duplicated from the input but manipulated to produce the result autoregressively \cite{DBLP:conf/acl/LewisLGGMLSZ20}, which is exactly the case here. The input to the model is the current question with previous conversational turns concatenated with a delimiter and the output consists of the structured representations. Once the input has been encoded, we calculate the soft cosine similarity between the current question and the given history turns. Unlike regular cosine similarity (which would result in zero for vectors with no overlapping terms), soft cosine similarity considers word similarity as well. The resultant value for soft cosine similarity ranges from 0 to 1 where 0 is no match and 1 represents an exact match with the history turns. Soft cosine simiarity can be calculated as:
\begin{equation}
    Soft cosine(q, h)= \frac{\sum_{i,j}^{N} s_{i,j}\  q_i\  h_j}{\sqrt{\sum_{i,j}^{N} s_{i,j} q_i q_j}\sqrt{\sum_{i,j}^{i-1} s_{i,j} h_i h_j}}
\end{equation}
where $q_i$ is the current question, $h_{j}^{N}$ represents history turns (j is equivalent to 0 and N is equivalent to i-1), and $s_{i,j} =similarity (feature_i, feature_j)$. 

We perform different experiments and conclude that the turns surpassing the threshold value of 0.75 contribute more in predicting the current answer span. Thus, all the turns that do not meet the threshold value would be filtered out. The resultant vectors are now passed on to the decoder to generate context and question entities for the respective turn. These entities are self-contained representations that capture the user's conversation flow from the previous history turns.

Once the SRs are generated for the previous history turns, the next step is to integrate them with the current question and the given context to provide some additional cues to answer the question. The architecture of CONVSR is shown in Fig.~\ref{sr}. The dotted line represents that both the current question and conversational history forms part of an input along with SRs to the ConvQA model.
%We use BERT-based \cite{DBLP:conf/naacl/DevlinCLT19} models not only to encode previous history along with the current question and given passage but also to encode the relationship between the SRs and the current question. BERT generates \textit{L} layers of hidden states for all tokens present in a given passage, question, previous history turns, and structured representations. 

\section{Experimental Setup}
\label{sec:setup}
In this section, we 
%discuss 
describe 
the experimental setup for our proposed model and the pipeline approach and compare our framework to the other state-of-the-art models.

\subsection{\textbf{Dataset Description}}
\subsubsection{\textbf{QuAC}}
Question Answering in Context (QuAC) \cite{choi-etal-2018-quac} consists of 100k question-answer pairs in a teacher-student information-seeking setting. The student seeks information on a topic provided with some background information, and the teacher attempts to satisfy the student's information need by engaging into a conversation. Since the test set is not made publicly available, %therefore, 
we randomly distribute 5\% of conversational dialogues in the training set following the strategy described in \cite{kim-etal-2021-learn}. We, then, utilize the distributed chunk as our validation set and report the test results.

\subsubsection{\textbf{CANARD}}
CANARD \cite{elgohary-etal-2019-unpack}, a dataset based on QuAC, consists of 40k question-answer pairs. The main idea behind CANARD is to convert the context-dependent questions of QuAC into context-independent or self-contained questions. These rewritten questions have the similar answer as that of the original questions. We utilize the training and development sets for training and validating the QR model, and the test set for evaluating the ConvQA models.

\subsection{\textbf{Training and Finetuning}}
To train the question understanding and entity generation modules, we have followed the technique of distantly supervised labeling introduced in \cite{DBLP:conf/sigir/ChristmannRW22}. The idea behind the technique is based on an intuition that if a piece of information (either entity or context) is essential for interpreting the follow-up question and has been omitted implicitly by the user, then it should be added in the completed version of the question. Based on this idea, the data for training is generated. We start with the complete questions and gather all the context and question entity mentions from it. For the incomplete or ambiguous follow-up questions, we keep on adding these entities to fill in the missing information. The entities are considered to be relevant for the incomplete question if an answer span is retrieved by adding them.

For training and evaluating the QR model, we use a publicly available dataset, CANARD \cite{elgohary-etal-2019-unpack} following the strategies discussed in \cite{vakulenko2021question, DBLP:journals/corr/abs-2004-01909}. The ConvQA models are trained on QuAC \cite{choi-etal-2018-quac} dataset with Adam optimizer with a learning rate of \textit{3e-5}.

\subsection{\textbf{ConvQA Models}}
Since both pipeline and CONVSR are model-agnostic, any ConvQA model can be utilized in the framework. The chosen models are widely utilized for comparison and have been proven to be performing well in ConvQA setting. We test the same models in both approaches to have a fair evaluation:
\begin{itemize}
\item BERT \cite{DBLP:conf/naacl/DevlinCLT19}: BERT is a pre-trained contextualized word representation model that is known to have empirically powerful results on different natural language tasks. BERT also works well on ConvQA datasets, although it was not designed for the task of ConvQA. It receives the context passage, current question, and conversational history as input.

    \item BERT-HAE \cite{10.1145/3331184.3331341}: BERT-HAE is based on BERT and introduces the idea of history answer embeddings to model the conversational history the concept  to model the conversational history. These contextualized history answer embeddings encode the answer tokens from the previous conversational turn into the model.
    
    \item RoBERTa \cite{DBLP:journals/corr/abs-1907-11692}: BERT is improved using advanced pre-training strategies to get the robustly optimized weights on huge corpora and the model is named as RoBERTa. It takes the same input as BERT unless stated otherwise. 
\end{itemize}

Apart from  evaluating the above models with dynamic history selection, we also experiment with the traditional ConvQA setting where the history turns with no selection criteria whatsoever, are appended to the current question. 

%michael: only one item, no need to itemize. 
%\begin{itemize}
%    \item \textbf{Prepending history turns:} 

Prepending previous conversational turns to the current question and the given context is still considered a simple yet very efficacious baseline in almost all ConvQA tasks. Hence, we experiment with the same here as well. Within prepending the conversational turns, we further investigate the effect of prepending only the initial turn (\textit{prepend init}), prepending only the last turn (\textit{prepend prev}), prepending initial and last history turns (\textit{prepend init + prev}), and prepending all the history turns (\textit{prepend all)}. For all these other experiments, we leverage RoBERTa \cite{DBLP:journals/corr/abs-1907-11692} to be the base model and adapt it according to the need of the task. The reason for choosing RoBERTa as a base model is that it is a top-performing model on the leaderboard of different conversational datasets and has shown its effectiveness in the ConvQA domain.
 %  \end{itemize}
   
\subsection{\textbf{Evaluation Metrics}}
For evaluation purpose, we follow the metrics used in \cite{choi-etal-2018-quac} to assess the performance of the models on the QuAC and CANARD datasets. The metrics include not only the F1 score but also the human equivalence score for questions (HEQ-Q) and dialogues (HEQ-D). HEQ-Q is the measure of the model's performance in retrieving the more accurate (or at least similar) answers for a given question. HEQ-D represents the same performance measure but instead of a question, it evaluates the overall dialog. 
%michael: to be self contained, please give the details (including definition) of the metrics we use in the experiments. 

\section{Results and Analysis}
\label{result}
We  conduct experiments on CONVSR and the competing baselines %on 
using the QuAC and CANARD 
datasets and report the results in this section.

\subsection{\textbf{CONVSR is viable for addressing incomplete questions in ConvQA}}
The first and foremost takeaway from the experiments is that our model works well in the ConvQA setting. The experiments are particularly designed to tackle the problem of incomplete or ambiguous questions. Instead of re-writing the questions to fill in the missing gaps in the given question, our model generates intermediate representations based on context and question entities. These entities  aid the answering process by providing cues to interpret the questions. From Table~\ref{result1}, we can clearly see that CONVSR consistently improves the model on both datasets. \begin{table*}[tbh!]
    \caption{Performance evaluation of the pipeline approach and our model using the  QuAC and CANARD datasets. The best scores are highlighted in bold}\begin{center}
\resizebox{0.8\linewidth}{!}{
    \begin{tabular}{p{2.0cm}|p{2.5cm}|p{1.8cm}|p{1.8cm}|p{1.8cm}} \hline
    \hline
    
\multicolumn{1}{c}{\textbf{Models}} & \multicolumn{1}{c}{\textbf{Approach}} & \multicolumn{1}{c}{\textbf{F1}} &  \multicolumn{1}{c}{\textbf{HEQ-Q}} &  \multicolumn{1}{c}{\textbf{HEQ-D}}\\ \hline
   \hline
    BERT & \begin{tabular}{l}Pipeline \\Ours\end{tabular} & \begin{tabular}{l}61.4 \\\textbf{62.7} \textcolor{blue}{\textbf{(+1.3)}}\end{tabular} & \begin{tabular}{l}57.4 \\\textbf{59.2} \textcolor{blue}{\textbf{(+1.8)}}\end{tabular}& \begin{tabular}{l}5.3 \\\textbf{6.2} \textcolor{blue}{\textbf{(+0.9)}} \end{tabular} \\
    \hline
   BERT-HAE  & \begin{tabular}{l}Pipeline \\Ours\end{tabular} & \begin{tabular}{l}61.5 \\\textbf{63.6} \textcolor{blue}{\textbf{(+2.1)}}\end{tabular} & \begin{tabular}{l}57.1 \\\textbf{59.3} \textcolor{blue}{\textbf{(+2.2)}}\end{tabular} & \begin{tabular}{l}6.0 \\\textbf{6.1} \textcolor{blue}{\textbf{(+0.1)}}\end{tabular}\\ 
   \hline
    RoBERTa  & \begin{tabular}{l}Pipeline \\Ours\end{tabular} & \begin{tabular}{l}66.1 \\\textbf{67.9} \textcolor{blue}{\textbf{(+1.8)}}\end{tabular}&\begin{tabular}{l}61.2\\\textbf{65.1} \textcolor{blue}{\textbf{(+3.9)}}\end{tabular} & \begin{tabular}{l}7.2\\\textbf{9.2} \textcolor{blue}{\textbf{(+2.0)}}\end{tabular}\\
    \hline

    \end{tabular}}
    \end{center}
    \label{result1}
    \vspace{-2em}
    
\end{table*}
\begin{table*}[htb!]
    \caption{The evaluation results based on traditional prepend baselines with SRs. The framework utilizes RoBERTa based model to generate the answers. The best scores are highlighted in bold.}
\begin{center}
    \resizebox{0.8\linewidth}{!}{
    \begin{tabular}{p{3.8cm}|p{2.2cm}|p{1.4cm}|p{1.4cm}|p{1.4cm}} \hline
    \hline
    
\multicolumn{1}{c}{\textbf{Models}} & \multicolumn{1}{c}{\textbf{Approach}} & \multicolumn{1}{c}{\textbf{F1}} &  \multicolumn{1}{c}{\textbf{HEQ-Q}} &  \multicolumn{1}{c}{\textbf{HEQ-D}}\\ \hline
   \hline
    PREPEND INIT + SR & \begin{tabular}{l}Pipeline \\Ours\end{tabular} & \begin{tabular}{l}59.4 \\\textbf{60.2} \textcolor{blue}{\textbf{(+0.8)}}\end{tabular} & \begin{tabular}{l}57.5 \\\textbf{58.7} \textcolor{blue}{\textbf{(+1.2)}}\end{tabular}& \begin{tabular}{l}4.7 \\\textbf{4.9} \textcolor{blue}{\textbf{(+0.2)}}\end{tabular} \\
    \hline
   PREPEND PREV + SR & \begin{tabular}{l}Pipeline \\Ours\end{tabular} & \begin{tabular}{l}62.2\\\textbf{64.4} \textcolor{blue}{\textbf{(+2.2)}}\end{tabular} & \begin{tabular}{l}60.1 \\\textbf{63.0} \textcolor{blue}{\textbf{(+2.9)}}\end{tabular} & \begin{tabular}{l}6.0\\\textbf{7.2} \textcolor{blue}{\textbf{(+1.2)}}\end{tabular}\\ 
   \hline
    PREPEND INIT + PREV + SR & \begin{tabular}{l}Pipeline \\Ours\end{tabular} & \begin{tabular}{l}60.1 \\\textbf{61.9} \textcolor{blue}{\textbf{(+1.8)}}\end{tabular} & \begin{tabular}{l}57.9 \\\textbf{59.3} \textcolor{blue}{\textbf{(+1.4)}}\end{tabular} & \begin{tabular}{l}5.8 \\\textbf{6.0} \textcolor{blue}{\textbf{(+0.2)}}\end{tabular}\\
    \hline
     PREPEND ALL & \begin{tabular}{l}Pipeline \\Ours\end{tabular} & \begin{tabular}{l}61.0 \\\textbf{62.4} \textcolor{blue}{\textbf{(+1.4)}}\end{tabular} & \begin{tabular}{l}58.1 \\\textbf{60.2} \textcolor{blue}{\textbf{(+2.1)}}\end{tabular} & \begin{tabular}{l}6.2 \\\textbf{6.6} \textcolor{blue}{\textbf{(+0.4)}}\end{tabular}\\
    \hline
  
\end{tabular}}
    \end{center}
    \label{result2}
    \vspace{-2em}
\end{table*}
\begin{table*}[htb!]
    \caption{The evaluation results are based on traditional prepend baselines without SRs.The framework utilizes RoBERTa based model to generate the answers. The best scores are highlighted in bold.}
    
    \begin{center}
\resizebox{0.8\linewidth}{!}{
     \begin{tabular}{p{3.8cm}|p{2.2cm}|p{1.5cm}|p{1.49cm}|p{1.4cm}} \hline
    \hline
    
\multicolumn{1}{c}{\textbf{Models}} & \multicolumn{1}{c}{\textbf{Approach}} & \multicolumn{1}{c}{\textbf{F1}} &  \multicolumn{1}{c}{\textbf{HEQ-Q}} &  \multicolumn{1}{c}{\textbf{HEQ-D}}\\ \hline
   \hline
    PREPEND INIT w/o SR & \begin{tabular}{l}Pipeline \\Ours\end{tabular} & \begin{tabular}{l}55.4 \\\textbf{58.9} \textcolor{blue}{\textbf{(+3.4)}}\end{tabular} & \begin{tabular}{l}54.5 \\\textbf{56.7} \textcolor{blue}{\textbf{(+2.2)}}\end{tabular}& \begin{tabular}{l}4.4 \\\textbf{4.8} \textcolor{blue}{\textbf{(+0.4)}}\end{tabular} \\
    \hline
   PREPEND PREV w/o SR & \begin{tabular}{l}Pipeline \\Ours\end{tabular} & \begin{tabular}{l}58.3\\\textbf{60.1}  \textcolor{blue}{\textbf{(+1.8)}} \end{tabular} & \begin{tabular}{l}56.4 \\\textbf{57.2}  \textcolor{blue}{\textbf{(+0.8)}}\end{tabular} & \begin{tabular}{l}\textbf{6.2}\\6.0  \textcolor{red}{\textbf{(-0.2)}}\end{tabular}\\ 
   \hline
    PREPEND INIT + PREV w/o SR  & \begin{tabular}{l}Pipeline \\Ours\end{tabular} & \begin{tabular}{l}57.9 \\\textbf{60.2}  \textcolor{blue}{\textbf{(+2.4)}}\end{tabular} & \begin{tabular}{l}55.5 \\\textbf{58.4}  \textcolor{blue}{\textbf{(+2.9)}}\end{tabular} & \begin{tabular}{l}5.4 \\\textbf{5.9}  \textcolor{blue}{\textbf{(+0.5)}}\end{tabular}\\
    \hline
    PREPEND ALL & \begin{tabular}{l}Pipeline \\Ours\end{tabular} & \begin{tabular}{l}58.0 \\\textbf{60.0} \textcolor{blue}{\textbf{(+2.0)}}\end{tabular} & \begin{tabular}{l}58.7 \\\textbf{59.4} \textcolor{blue}{\textbf{(+0.7)}}\end{tabular} & \begin{tabular}{l}6.1 \\\textbf{6.4} \textcolor{blue}{\textbf{(+0.3}}\end{tabular}\\
    \hline
  
\end{tabular}}
    \end{center}
    \label{result3}
\end{table*}

\subsection{\textbf{CONVSR outperforms all the traditional baselines}}
We observe from Table~\ref{result2} that generating SRs yields better results even in the traditional prepend baselines. Out of all the variations, \textit{prepend prev} provides the highest F1 score. It confirms the intuition that incomplete questions usually take the context and entities of the last question asked to fill in the missing information gap. \textit{Prepend init} results in a low F1 score mainly because of the reason that the flow of the conversation keeps on changing. The first question asked in the conversation does not necessarily provide  the related context and question entities to the current question. Table~\ref{result3} shows the accuracy scores without utilizing SRs in traditional prepend baselines. Comparing the two tables, we can clearly see that SRs provide an edge to the model in predicting correct answer spans.

\subsection{\textbf{Role of slots in SR}}
The two slots in SRs play a vital role in understanding an incomplete question. Table~\ref{result4} shows a comparison of the F1 score when the slots are omitted on purpose one by one. The role of question entities is crucial. Skipping question entities from a question results in a major decline in the F1 score. 

\begin{table}[h!]
    \caption{Effect of SR slots on accuracy score.}
\begin{center}
\
    \begin{tabular}{p{3.0cm}|p{2.5cm}} \hline
    \hline
\multicolumn{1}{c}{\textbf{Models}} & \multicolumn{1}{c}{\textbf{F1}}\\ \hline
   \hline
    CONVSR & \textbf{67.9}  \\
    \hline
   w/o context entity & 64.3 \\ 
   \hline
    w/o question entity & 62.8 \\ 
   \hline
  
    \end{tabular}
    
    \end{center}
    \label{result4}
\vspace{-2.5em}
\end{table}

\subsection{\textbf{Verbose questions lead to decline in F1 score}}
The results in Table~\ref{result5} show that question re-writing results in lengthy questions, which may cause losing valuable cues from the conversation flow, hence, the decline in results. Also, QR results in more proper nouns, which shows that generating QRs requires mapping more entities within the given context. This mapping adds more complexity in generating questions from the scratch. Consequently, this may also result in a decline in the F1 score.

\begin{table}[htb!]
\caption{The effect of length on the accuracy score of predicted answer.}
\begin{center}
\resizebox{1\linewidth}{!}{
    \begin{tabular}{p{2.5cm}|p{0.5cm}|p{0.5cm}|p{0.5cm}|p{0.5cm}} \hline
    \hline
\multicolumn{1}{c}{\textbf{Methods}} & \multicolumn{1}{c}{\textbf{Avg Length}} &  \multicolumn{1}{c}{\textbf{Pronoun}} &  \multicolumn{1}{c}{\textbf{Proper Noun}} &
\multicolumn{1}{c}{\textbf{F1}}\\ \hline
   \hline
    Original + SR & 5.5  & 0.5 & 1 & 67.9  \\
\hline
   Question Rewriting & 9 & 0 & 2.5 & 66.1 \\ 
\hline  
    \end{tabular}}
    \end{center}
    \label{result5}
    \vspace{-2em}
\end{table}

\section{Conclusion and Future Work}
\label{sec:conclusion}

In this paper, we have argued that generating the paraphrases of incomplete and ambiguous questions can take out questions from the conversational context, thereby impeding the underlying essence of conversational question answering (ConvQA). Moreover, the rewritten questions are lengthy and verbose and, thus, add complexity to the answer retrieval part. In an attempt to overcome these issues, we 
%introduced 
have proposed CONVSR, a conversational question answering model which utilizes structured representations in the form of both context entity and question entity for predicting the answer span. Our experimental results demonstrate the significance of structured representation (SR) generation within a ConvQA setting. Our 
%framework 
model significantly improves ConvQA performance on both QuAC and CANARD datasets, i.e., as compared to the existing state of the art. Our approach leverages strategies from different research fields and their strategic paradigms. The idea of generating intent-explicit SRs is taken from symbolic AI, whereas, the tasks of question rewriting and question answering have their roots embedded in the IR community. 

One of the promising directions for our future work involves generating more context-aware SRs that can be utilized on the heterogeneous sources of text-based ConvQA. Furthermore, we %intend 
plan to scale up our proposed 
%framework 
model to target the open-domain ConvQA setting. 
\bibliographystyle{IEEEtran}
\bibliography{IJCNN}

\end{document}